# A DYNAMIC VULNERABILITY MAP TO ASSESS THE RISK OF ROAD NETWORK TRAFFIC UTILIZATION

*Michel Nabaa[1], Cyrille Bertelle[1], Antoine Dutot[1], Damien Olivier[1] and Pascal Mallet[2]*

[1]*LITIS laboratory, 25 rue Phillipe Lebon, F-76085 Le Havre*
Email: {Michel.Nabaa, Cyrille.Bertelle , Antoine.Dutot ,Damien.Olivier}@univ-lehavre.fr

[2] *Le Havre agglomeration, 19 rue Georges Braque, F-76600 Le Havre.*
Email: Pascal.Mallet@agglo-havraise.fr

**Abstract**. *Le Havre agglomeration (CODAH) includes 16 establishments classified Seveso[3] with high threshold. In the literature, we construct vulnerability maps to help decision makers assess the risk. Such approaches remain static and do take into account the population displacement in the estimation of the vulnerability. We propose a decision making tool based on a dynamic vulnerability map to evaluate the difficulty of evacuation in the different sectors of CODAH. We use a Geographic Information system (GIS) to visualize the map which evolves with the road traffic state through a detection of communities in large graphs algorithm.*

## 1 INTRODUCTION

The population of the Seine estuary is exposed to several types of natural and industrial hazards. It is included in the drainage basin of the "*Lézarde*" and is also exposed to significant technological risks. The modeling and assessment of the danger is useful when it intersects with the exposed stakes. The most important factor is people. Recent events have shown that our agglomerations are vulnerable in front of emergency situations. The examination of impacted populations remains a difficult exercise. In this context, the Major Risk Management Direction team (DIRM) of Le Havre Agglomeration (CODAH) has developed a model of spatial and temporal population exposed allocation PRET-RESSE; the scale is the building (Bourcier and Mallet 2006); thus by distinguishing their day and night occupation. The model was able to locate people during the day both in their workplace and their residence (the unemployed and retirees). Although the model was able to locate the diurnal and nocturnal population, it remains static because it does not take into account the daily movement of people and the road network utilization.

For a better evacuation of people in the case of a major risk, we need to know the state of road traffic to determine how to allocate the vehicles on the road network and model the movement of these vehicles. In fact, the panic effect of some people can lead to accidents and traffic jams, which may be too grievous with a danger that spreads quickly. The panic generally results from the lack of coordination and dialogue between individuals (Provitolo, 2007).

In the literature, several models were developed to calculate a score of the vulnerability related to the road network utilization. This score may depend on social, biophysical, demographical or other aspects. Most of these models adopt a pessimistic approach to calculate this vulnerability: this case is met when a group of people in a hazardous area decide all to take the same route to evacuate this area, which unfortunately happens quite often in the real world evacuation situations. Although it helps decision-makers to estimate the risk by a census vulnerability map, this approach remains static and does not take into account the evolution of the road network traffic.

In this paper, we have to simplify the representation of the population displacement, which is a complex phenomena. We also propose a dynamic and pessimistic approach related to the access to the road network. To this end, we model the road network by a dynamic graph (the dynamics is due to the traffic evolution). A simple model based on traffic flow will also be proposed. Then, we apply a self-organization algorithm to detect communities on the graph belonging to the collective intelligence algorithms. The algorithm allows us to define the different vulnerable neighborhoods of the agglomeration in the case of an evacuation due to a potential danger, while taking into account the evolution of the road network traffic. The result of this algorithm will be visualized into a GIS on a dynamic vulnerability map which categorizes various sectors depending on the difficulty of access to the road network. The map will help decision-makers in a better estimation of risk in the communes of the CODAH. It will enrich PRET-RESS static model developed at the CODAH, taking into account the mobility of the population.

---

[3] Directive SEVESO is an European directive, it lays down to the states to identify potential dangerous site. It intends to prevent major accidents involving dangerous substances and limit their consequences for man and the environment, with a view to ensuring high levels of protection throughout the Community.

## 2 VULNERABILITY ASSESSMENT APPROACHES

Traditional methods of conception and evaluation of the population at risk do not sometimes treat the behavioral evacuee's response (e.g. initial response to an evacuation, travel speed, family interactions / group, and so on.); they describe prescriptive rules as the travel distance. These traditional methods are not very sensitive to human behavior for different emergency scenarios. The computerized models offer the potential to evaluate the evacuation of a neighborhood in emergency situations and overcome these limitations (Castel 2006).

Recently, some interesting applications have been developed by including the population dynamics, the models of urban growth patterns and land use.

For computer modelers, this integration provides the ability to have computing entities as agents that are linked to real geographical locations. For GIS users, it provides the ability to model the emergence of phenomena by various interactions of agents in time and space by using a GIS (Najlis and North 2004).

Many researchers have emphasized the need to create models of based vector Multi Agent Systems (MAS), which may require topological data structure provided in GIS (Parker 2004). In this way, one can represent an ABM in which may coexist n levels of organizations with several classes of agents (e.g. Level 1: individuals or companies, Level 2 and 3: agents, cities, communities. . .) (Daude 2005).

In (Cutter et al 2000), the author presents a method to spatially estimate the vulnerability and treats the biophysical and social aspects (access to resources, people with evacuation special needs, people with reduced mobility ...).

Several layers are created into the GIS (a layer by a danger), and all these layers are combined into one composed of intersecting polygons to build a generic vulnerability map. To complete, it was necessary to take into account the infrastructure and various possible emergency exits. So, a new map has been constructed and a new layer has been incorporated. This work has been applied to the George Town canton in which we find various natural and industrial risks, and where there are different types of people.

In the neighborhood evacuation cases on a micro scale, a number of studies based on micro simulation have been developed. In their paper (Church and Cova 2000), the authors presented a model to estimate the necessary time to evacuate a neighborhood according to the effective of the population, number of vehicles, roads capacity and number of vehicles per minute. The model is based on the optimization in order to find the critical area around a point at a potential danger in a pessimistic way. This model has been coupled with a GIS (ArcInfo) to visualize the results (identify evacuation plans) and construct an evacuation vulnerability map for the city (Santa Barbara).

Cova and Church (1997) opened the way on the study based on geographic information systems to evacuate people. Their study identified the communities that may face transport difficulties during an evacuation. Research has modeled the population by lane occupation during an evacuation emergency using the city of Santa Barbara.

An optimization based model (graph partitioning problem) was realized to find the neighborhood that causes the highest vulnerability around each node in the graph and a vulnerability map around nodes in the city was constructed. A constructive heuristic has been used to calculate the best cluster around each node. This heuristic was developed in C and the result was displayed on a map (with ARCINFO).

Nevertheless, in this approach, we predefine the maximum number of nodes in a neighborhood, which may not always be realistic and does not take into account the traffic evolution during the calculation of critical neighborhoods. So, the vulnerable neighborhoods don't evolve according to traffic state.

In our work, we try to build a dynamic vulnerability map evolving with the traffic dynamics, in which the nodes number of in a critical neighborhood, is not predetermined and can change depending on traffic state.

## 3 PROBLEMATIC

In this paper, the vulnerability is related to the access to the road network. We are persuaded that an accident on the network may cause traffic jams and therefore serious problems. So, it is important to have a pessimistic approach which takes into account the worst behavior of evacuees during a disaster (e.g. all individuals evacuate by taking the same exit route).

To address this vulnerability, we have to finely represent the population and the dynamic state of road traffic. In PRET-RESSE model developed within the major risks management team of CODAH, we have ventilated the day / night population at the buildings. The model was able to locate people during the day both in their workplace and their residence (the unemployed and retirees). It has been estimated that people will be in their residence during the night. A displacement survey was also realized in CODAH agglomeration and will be included in our work. PRET-RESS will be enriched by our model to dynamically assess the vulnerability related to the road traffic evolution.

# 4 DYNAMIC MODEL

## 4.1 System architecture

The system consists of two related modules. The first one is dedicated to simulate the flow and apply the communities detection algorithm on the graph. The second one allows visualizing the result into a GIS. The architecture is illustrated in the following figure.

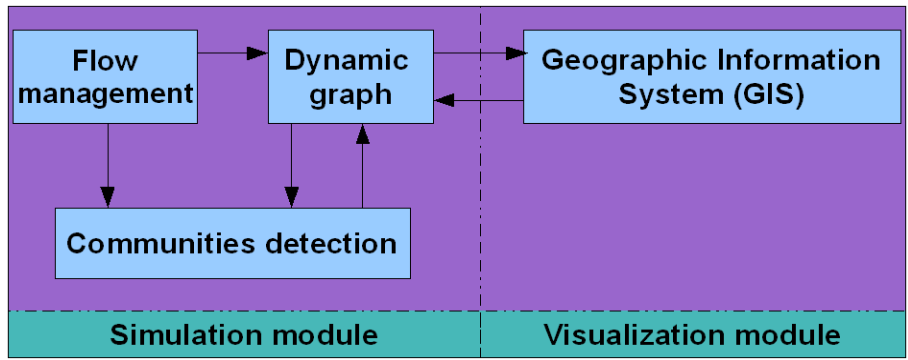

**Figure 1**: System architecture

The simulation module contains three components:
- The dynamic graph extracted from the road network layer and detailed in the following section.
- The flow management component consists of vehicles flow simulator applied on the graph.
- The communities' detection component, detailed in the section 5. Its takes the extracted graph and the current flow as an input and returns the formed communities according to the current state of road traffic.
The visualization module consists of the road network layer integrated into the GIS. This module communicates with the simulation module: The graph is constructed from this module, which in turn get the simulation result and visualize it as a dynamic vulnerability map.

## 4.2 Environment modeling

The road network is integrated as a layer in the Geographic Information System (GIS). From this layer, we extract the data by using the open source java GIS toolkit Geotools. This toolkit provides several methods to manipulate geospatial data and implements Open Geospatial Consortium (OGC) specifications, so we can read and write ESRI shapefile format. Once the necessary road network data extracted, we use the GraphStream tool developed within LITIS laboratory of Le Havre to construct a graph. This tool is designed for modeling; processing and visualizing graphs.
The data extracted from network layer contains the roads circulation direction, roads id, roads type, their lengths and geometry.
The extracted multigraph $G(t) = (V(t), E(t))$ represents the road network where $V(t)$ is the set of nodes and $E(t)$ the set of arcs. We deal with a multigraph because there was sometimes more than one oriented arc in the same direction between two adjacent nodes due to multiple routes between two points in Le Havre road network. GraphStream facilitates this task because it is adapted to model and visualize multigraphs. In the constructed multigraph:
- The nodes represent roads intersections;
- The arcs represent the roads taken by vehicles;
- The weight on each arc represents the needed time to cross this arc, depending on the current load of the traffic;
- Dynamic aspect relates to the weights of the arcs, which can evolve in time, according to the evolution of the fluidity of circulation;
we have also constructed a Voronoi tessellation (Thiessen polygon) around nodes and projected the population in buildings on these nodes. The population in buildings is extracted from PRET-RESS model.

## 4.3 Traffic management

For a better evacuation of people in a major risk situation, we need to know the state of the road network to determine how to allocate the vehicles on this network and to model the movement of these vehicles. Different types of models can be adopted:

- The microscopic model details the behavior of each individual vehicle by representing interactions (modeled by a car following model) with other vehicles and in general by using a spatialization. We can extend the model by adding a regulation model with priorities rules, traffic lights… Microscopic models may be applied on crowds movement as in boids collective approaches (Reynolds 1987). In our problem, a microscopic model is used on the scale of a sector or a neighborhood. It has the advantage to model vehicle behavior in an evacuation of a neighborhood, people panic, interactions between vehicles, accidents…
-The macroscopic model is based on the analogy between vehicular traffic and the fluid flow within a canal. It allows us to visualize the flow on the roads rather than individual vehicles. It is used at many sectors or the entire city scale.
The hybrid model allows coupling the two types of dynamics flow models within the same simulation. Several works have already borrowed this direction (Hennecke et al. 2000, Bourrel and Henn 2002, Magne et al. 2000). However, this approach is relatively new and very few have adopted it to our knowledge (Hman et al. 2006).
In risk context, the use of a hybrid model is very important especially when dealing with large volume of data: changing the scale from micro to macro in a region where we haven't a crisis situation (everything is normal) allows to economize the computation and the change from macro to micro in a critical situation allows to zoom and detect the behaviors and interactions between entities in danger.
In this paper, we used a simple model of macroscopic flow:
- A set of flow of cars moving from one arc to another adjacent one.
- The arcs are limited in capacity of vehicles.
- The flow can be broken and two or more flows can gather on a node.
- Traffic jams may appear in certain places of the road network; those places will be more vulnerable than others.
We have adopted a macroscopic model in which flows circulate normally (no accidents) because the goal now is to establish a dynamic pessimistic vulnerability map which is not always the case in the real world (90% of people takes an exit route and the rest takes another route for example). Hence, it is important to have in the near future a micro approach with a change of scale (from micro to macro and vice versa during the simulation) to simulate scenarios of danger in real time (accidents, behavior of drivers, vehicles interactions ...), a study on which we are working actually .

## 4.4 Complex system

A complex system is characterized by numerous entities of the same or different nature that interact in a non-trivial way (non-linear, feedback loop ...); the global emergence of new properties not seen in these entities: Properties or evolution cannot be predicted by simple calculations.
CODAH can be seen as a complex system in which the environment may influence on evacuation by imposing some rules which may reduce the flow fluidity (existence or not of safe refuge and emergency exits, routes traffic direction, priorities, traffic lights...) and vice versa (a fire or an accident may cause damages and change the environment). This system is in perpetual evolution; it is far from equilibrium dynamics with an absence of any global control. Some organizations may appear or disappear according to different interactions. Entities as vehicles and pedestrians interact between them and with environment.

## 5 COMUNITIES DETECTION ALGORITHM

Our aim is to identify communities in graphs, i.e. dense areas strongly linked to each other and more weakly linked to the outside world.
The concept of communities in a graph is difficult to explain formally. It can be seen as a set of nodes whose internal connections density is higher than the outside density without defining formal threshold (Pons 2005). The goal is to find a partition of nodes in communities according to a certain predefined criteria without fixing the number of such communities or the number of nodes in a community.
Radicchi (Radicchi et al., 2004) proposes two possible definitions to quantify a community definition:

• Community C in a Strong Sense: $d_C^{out}(i) > d_{\overline{C}}^{out}(i), \forall i \in C$. A node belongs to a strong community if it has more connections within the community than outside.

• Community C in a Weak Sense: $\sum_{i \in C} d_C^{out}(i) > \sum_{i \in C} d_{\overline{C}}^{out}(i), \forall i \in C$. A community is qualified as weak if the sum of all degrees inside is more important than the sum of degrees towards the rest of the graph.

$d_C^{out}(i)$ is the exiting edges number from a node $i$ belonging to the community $C$ towards the nodes of the same community.

$d_{\overline{C}}^{out}$ is the exiting edges number from a node $i$ belonging to the community $C$ towards the nodes of other communities.
Finding organizations is a new field of research (Albert and Barabasi, 2002; Newman, 2004a). Interesting works were developed in the literature on the detection of structure in large communities in graphs (Clauset et al. 2004, Newman

2004a; b, Pons and Latapy 2006). In our problem, we look for a self-organization in networks with an evolutionary algorithm close to the detection of communities in large graphs and belonging to collective intelligence algorithms. This algorithm is based on the spread of forces in graphs.

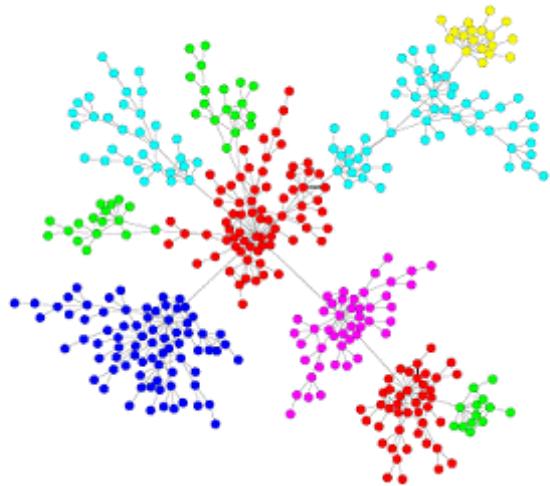

**Figure 2**: Communities detection in a graph

The algorithm principle is to color the graph using pheromones and it uses several colonies of ants, each of a distinct color. Each colony will collaborate to colonize zones, whereas colonies compete to maintain their own colored zone (see figure 2). Solutions will therefore emerge and be maintained by the ant behavior. The solutions will be the color of each vertex in the graph. Indeed, colored pheromones are deposited by ants on edges. The color of a vertex is obtained from the color having the largest proportion of pheromones on all incident edges.

We have an interaction between each two local adjacent nodes according to the attraction force that exists between them. This force depends in our case on *N/C*, where *N* is the number of vehicles on the arc between 2 nodes neighbors and *C* represents vehicles capacity of the arc. This report was chosen because, in each community, we will have a large number of vehicles which all decide to exit through the same route in the case of a potential danger; this responds well to one of the purposes listed in beginning to have a pessimistic approach in the calculation of vulnerability. The algorithm has the advantage of not allowing the breaking of a link between 2 adjacent nodes to maintain the structure of the road network. When the traffic evolves, the algorithm detects that and communities can change or disappear as a result of local forces that change between the nodes locally.

At each simulation time step, the flow on the arcs change following traffic conditions and the attraction forces may change also. Once communities are formed on the graph, the result will be transmitted to the road network layer into the GIS to be visualized.

## 6   CONCLUSION

In this paper, we have proposed a decision making tool to assess the danger depending on the road network use by the vehicles. This tool enables decision makers to visualize, on a geographic information system, a dynamic vulnerability map linked to the difficulty of evacuating the various streets in the metropolitan area of Le Havre agglomeration. We simulated the road network traffic by using a simple model of vehicles flow. A communities detection algorithm in the large graphs was adopted. It enabled us to form communities in a graph thanks to local force propagation rules between adjacent nodes. The communities evolve according to the current state of road network traffic. The result of the evolution of communities is visualized by using a GIS.

The adopted approach allowed us to estimate the risk due to the use of the road network by vehicles and categorize Le Havre agglomeration areas by their vulnerability. We will complete our work by using a micro model of traffic with a possibility of change from micro to macro and vice versa when necessary, and this depending on the situation.

## 7   REFERENCES

[1] Aaron Clauset, M. E. J. Newman and Cristopher Moore (2004). Finding community structure in very large networks. *Phys. Rev. E 70*, 066111.
[2] A. Hennecke, M. Treiber and D. Helbing (2000). Macroscopic simulation of open systems and micro-macro link. In M. Schreckenberg D. Helbing and H. J. Herrmann, editor, *Traffic and Granular Flow '99 : Social, Traffic, and Granular Dynamics*, pages 383–388. Springer.

[3] Castel, C.J.E. (2006). Developing a prototype agent-based pedestrian evacuation model to explore the evacuation of King's Cross St Pancras underground station, centre for advanced spatial analysis (university college London): working paper 108, London.

[4] Church, R.L. and Cova, T.J. (2000). Mapping evacuation risk on transportation networks using a spatial optimization model. *Transportation Research Part C: Emerging Technologies*, *8(1-6)*: 321-336.

[5] Cova, T.J. and Church, R.L. (1997) Modelling community evacuation vulnerability using GIS. *International Journal of Geographical Information Science*, *11(8)*: 763-784.

[6] Cutter, S.L., J.T. Mitchell and M.S. Scott (2000). Revealing the vulnerability of people and places: a case study of Georgetown County, South Carolina. *Annals of the Association of American Geographers 90 (4)*: 713-737.

[7] E. Bourrel and V. Henn (2002). Mixing micro and macro representation of traffic flow: a first theoretical step. *In Proceeding of the 9th Meeting of the Euro Working Group on transportation*.

[8] E. Daudé (2005). Systèmes multi-agents pour la simulation en géographie : vers une Géographie Artificielle, chapter 13, pages 355–382. *in Y. Guermont (dir.), Modélisation en Géographie : déterminismes et complexités*, Hermès, Paris.

[9] F. Radicchi, C. Castellano, F. Cecconi, V. Loreto and D. Parisi. Defining and identifying communities in networks. *In Proceedings of the National Academy of Sciences, volume 101*, pages 2658–2663, 2004.

[10] GraphStream : un outil de modélisation et de visualisation de graphes dynamiques. *Distribué sous licence libre (GPL)*.Http://graphstream.sourceforge.net.

[11] J.C. Bourcier and P. Mallet (2006). Allocation spatio-temporelle de la population exposée aux risques majeurs. Contribution à l'expologie sur le bassin de risques majeurs de l'estuaire de la seine: modèle PRET-RESSE. *Revue internationale de Géomatique, 16(10)* : 457-478.

[12] L. Magne, S. Rabut and J. F. Gabard (2000). Toward an hybrid macro and micro traffic flow simulation model. *In INFORMS spring 2000 meeting*.

[13] M. E. J. Newman (2004). Detecting community structure in networks, *Eur. Phys. J. B 38*, 321–330.

[14] M. E. J. Newman (2004). Fast algorithm for detecting community structure in networks. *Phys. Rev. E 69*, 066133.

[15] M.S. El Hman, H. Abouaïssa, D. Jolly and A. Benasser (2006). Simulation hybride de flux de trafic basée sur les systèmes multi-agents. *In 6e Conférence Francophone de MOdélisation et SIMulation - MOSIM'06*.

[16] Najlis, R. and M. J. North (2004). Repast for GIS. *Proceedings of Agent 2004: Social Dynamics: Interaction, Reflexivity and Emergence*, University of Chicago and Argonne National Laboratory, IL, USA.

[17] D. C. Parker (2004). Integration of geographic information systems and agent-based models of land use : Challenges and prospects. *In Maguire, D., J. M. F., Goodchild,and M., Batty, (Eds), GIS, Spatial Analysis and Modelling, Redlands, CA:* ESRI. Press.

[18] Pascal Pons (2005). Détection de structures de communautés dans les grands réseaux d'interactions. *Septièmes Rencontres Francophones sur les aspects Algorithmiques des Télécommunications*. Giens, France.

[19] Pascal Pons and Matthieu Latapy(2006). Computing communities in large networks using random walks. *Journal of Graph Algorithms and Applications. Vol. 10, no. 2,* pp. 191-218.

[20] D. Provitolo (2007). A proposition for a classification of the catastrophe systems based on complexity criteria. *In European Conference Complex Systems-EPNACS'07, Emergent Properties in Natural and Artificial Complex Systems, Dresden,* pages 93–106.

[21] Reynolds C. (1987). Flocks, Herds, and Schools*:* A Distributed Behavioral Model. Computer Graphic*s. SIGGRAPH 87 Conference, vol. 21(4),* 25–34.